\documentclass[accepted]{uai2026}

\usepackage[american]{babel}

\usepackage{natbib}
\usepackage{mathtools}
\usepackage{amsmath,amssymb,amsfonts}
\usepackage{booktabs}
\usepackage{tcolorbox}
\usepackage{algorithm2e}
\usepackage{enumitem}
\usepackage{graphicx}
\usepackage{microtype}
\usepackage{dblfloatfix}
\usepackage{placeins}

\makeatletter
\renewcommand{\maketitlehooka}{%
    \vbox to 2.375in\bgroup
    \noindent\textit{Preprint, Under Review}\par\vspace{0.1in}
}
\makeatother
 
\title{Don't Blink: Evidence Collapse during Multimodal Reasoning}

\author[1]{Suresh Raghu\textsuperscript{\textdagger}}
\author[1]{Satwik Pandey\textsuperscript{\textdagger}}
\affil[1]{Independent Researcher, New Delhi, India}
\affil[ ]{\texttt{sureshraghu0706@gmail.com, psatwik2711@gmail.com}}
 
\begin{document}

\maketitle
\begingroup
\renewcommand{\thefootnote}{\fnsymbol{footnote}}
\footnotetext[2]{Equal contribution.}
\endgroup
\begin{abstract}
Reasoning VLMs can become more accurate while progressively losing visual grounding as they think. This creates task-conditional danger zones where low-entropy predictions are confident but ungrounded, a failure mode text-only monitoring cannot detect. Evaluating three reasoning VLMs on MathVista, HallusionBench, and MMMU\_Pro, we find a pervasive evidence-collapse phenomenon: attention to annotated evidence regions drops substantially, often losing over half of evidence mass, as reasoning unfolds. Full-response entropy is the most reliable text-only uncertainty signal under cross-dataset transfer, yet adding vision features with a single global linear rule is brittle and often degrades transfer. An entropy-vision interaction model reveals a task-conditional regime: low-entropy, visually disengaged predictions are hazardous on sustained visual-reference tasks but benign on symbolic tasks. Using this structure, a targeted vision veto reduces selective risk by up to 1.9 percentage points at 90\% coverage, while avoiding degradations where disengagement is expected. The results support task-aware multimodal monitoring for safe deployment under distribution shift. All code is publicly available at \url{https://github.com/R-Suresh07/Dont-Blink}.
\end{abstract}

\section{Introduction}

Uncertainty quantification is essential for the safe deployment of AI systems in high-stakes applications. Vision-Language Models (VLMs) trained via reinforcement learning to produce explicit reasoning traces, rather than relying on prompt-elicited chain-of-thought (hereafter RVLMs) achieve state-of-the-art accuracy on multimodal benchmarks spanning mathematical reasoning, expert-level knowledge, and visual question answering. Yet accuracy alone is insufficient for reliable decision-making: a model correct 80\% of the time but reporting 95\% confidence on every prediction is dangerous in domains where confident errors carry severe consequences. Calibration, the alignment of predicted confidence with true correctness probability, is a well-known challenge in modern deep neural networks~\citep{guo2017calibration}, and evidence suggests it may worsen after reinforcement learning-based post-training.

\begin{figure*}[!t]
\centering
\includegraphics[width=0.95\textwidth]{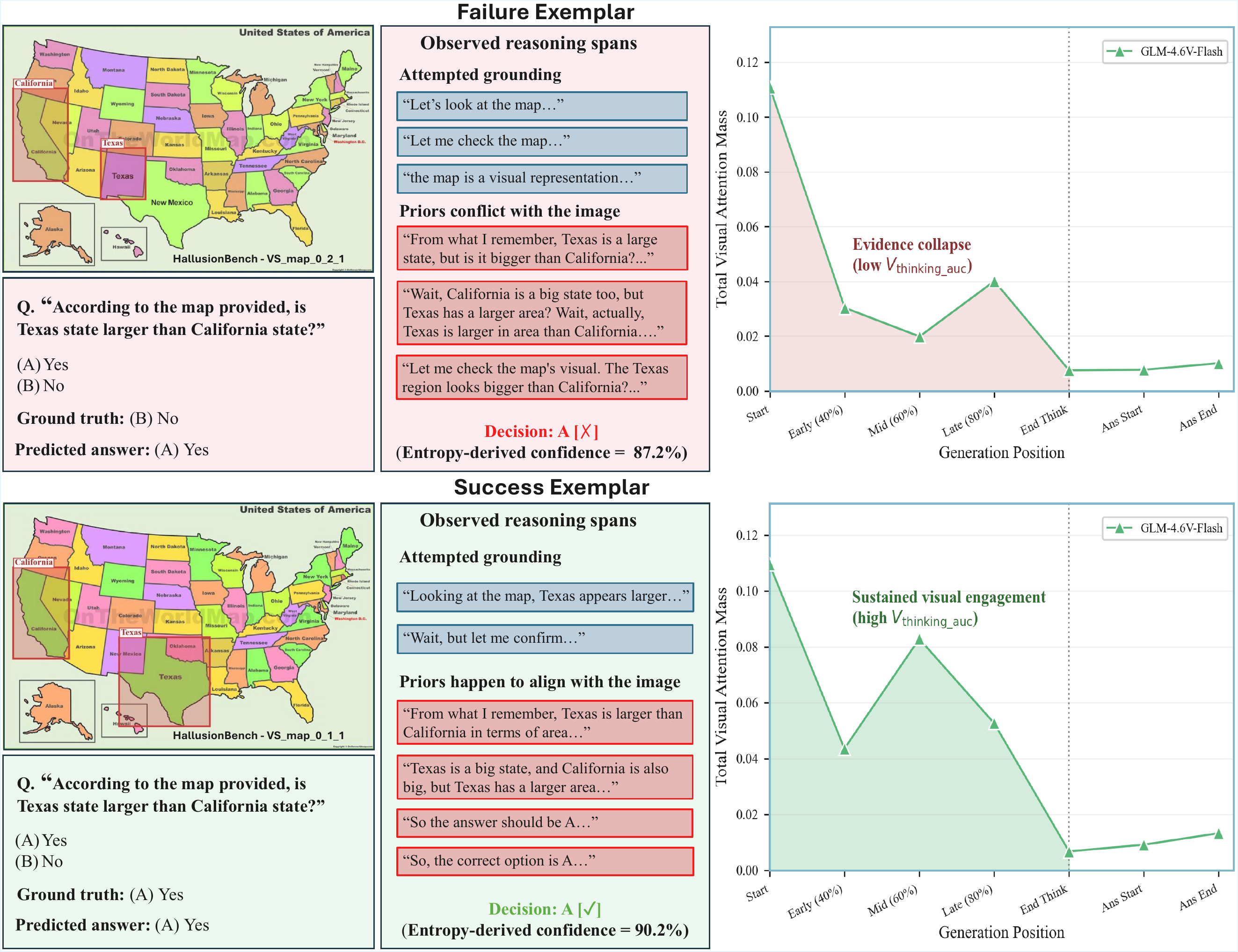}
\caption{\textbf{Evidence collapse creates confident errors invisible to text-only monitoring.} Two HallusionBench exemplars processed by GLM-4.6V-Flash on nearly identical map-comparison questions. \textbf{Top (Failure):} On an adversarially modified map, the model attempts visual grounding early but textual priors override image evidence mid-reasoning (red spans), producing a confident wrong answer with no attention recovery. \textbf{Bottom (Success):} On the unmodified map, textual priors align with the image and visual engagement is sustained, peaking at mid-generation. Entropy-derived confidence differs by only 3 percentage points (87.2\% vs.\ 90.2\%), yet the visual attention trajectories, summarized by $V_{\mathrm{thinking\_auc}}$, are qualitatively different. Text-only confidence cannot separate these cases; cumulative visual attention can.}\label{fig:overview}
\end{figure*}

Existing uncertainty methods for language models operate entirely over text token distributions: token entropy~\citep{xiong2024efficient}, semantic clustering~\citep{kuhn2023semantic}, prompt-based confidence elicitation~\citep{xiong2024can}, and entropy discrimination that can emerge from RL post-training~\citep{sharma2025think}. Full-response entropy, averaged over the entire generation, can be a strong uncertainty monitor, but it is structurally incapable of distinguishing whether a confident prediction is grounded in visual evidence or has drifted into text-prior-driven confabulation. This motivates examining signals from the model's own visual processing for failure modes that text-only monitoring cannot detect.

Work on visual evidence attribution shows that visual attention carries information beyond text signals~\citep{liu2025seeing}, but prior methods target intervention at single generation steps rather than trajectory-level measurement across extended reasoning.

We make four contributions:
\begin{enumerate}
\item \textbf{Trajectory-level measurement of visual grounding in RVLMs.} Evidence collapse is universal (9/9 cells) and often peaks mid-reasoning rather than at answer time.
\item \textbf{A strong transfer baseline for uncertainty monitoring.} Full-response entropy outperforms alternatives under cross-dataset transfer, but cannot detect visual disengagement.
\item \textbf{Discovery of a task-conditional failure regime.} An entropy--vision interaction model reveals a ``confident-but-blind'' penalty on sustained visual-reference tasks, absent or inverted on symbolic tasks.
\item \textbf{Task-aware monitoring design principles.} A targeted vision veto validates the task-conditional structure: it helps where the interaction predicts danger and hurts where it predicts safety.
\end{enumerate}

\section{Related Work}
\label{sec:related}

\textbf{Uncertainty estimation in language models.}
Modern deep neural networks are poorly calibrated despite high accuracy~\citep{guo2017calibration,naeini2015obtaining,nixon2019measuring}. Text-only uncertainty methods derive confidence from output distributions: token-level entropy and single-sample estimators~\citep{xiong2024efficient}, semantic uncertainty via clustering sampled outputs under linguistic invariances~\citep{kuhn2023semantic}, and prompt-based confidence elicitation~\citep{xiong2024can}; sequence-level entropy discrimination can emerge after RL post-training~\citep{sharma2025think}. These methods assume a unimodal text setting and do not account for visual evidence competing with accumulating text priors. Work on VLM uncertainty extends this line. \citet{chen2024unveiling} and \citet{he2023investigating} examine alignment and fine-tuning effects on calibration, while recent benchmarking reveals that VLM uncertainty scores are weak and model-dependent~\citep{wang2026vlmuqbench,lu2025prolonged,ortiz2026abstention}. Yet even these multimodal studies derive confidence from text outputs rather than modality-specific grounding dynamics. Across both lines, no method incorporates modality-specific grounding signals, and none tracks how such signals evolve across the generation trajectory.

\textbf{Visual evidence attribution.}
Attention-based interpretability tools, from rollout methods to gradient-weighted attribution, aim to localize which image regions influence VLM outputs. \citet{liu2025seeing} introduced evidence attribution for VLMs, showing that models can attend to correct evidence yet still err, but analyzed instruct-style generations at single endpoints and per-layer probes rather than reasoning trajectories. Recent intervention-focused work addresses grounding drift: \citet{jiang2026kvsmooth} frame hallucination as semantic drift during decoding and propose KV-cache smoothing; \citet{adavboost2026} propose visual grounding entropy to estimate hallucination risk at individual generation steps. External verification approaches predict trust for spatial reasoning via object detection and geometric checks~\citep{imran2026trust}, but operate outside the model's internal representations.
No prior work measures grounding trajectories across the full reasoning chain of RVLMs or connects trajectory dynamics to correctness under task-conditional variation.

\section{Motivation}
\label{sec:motivation}

Entropy discrimination, where correct generations exhibit lower output entropy than incorrect ones, can emerge from RL post-training and is attractive for reasoning models because it requires no additional labels. This signal is strong but structurally compromised in the multimodal setting, motivating a modality-specific alternative.

RVLMs reason over \emph{fixed} visual tokens: the image is encoded once and does not change as the reasoning chain grows token by token. As text accumulates, attention to the growing text context competes with attention to the fixed visual context, creating systematic pressure for visual attention to decay over the course of generation. Entropy operates entirely over text token distributions and cannot detect this pressure. A model answering a diagram-dependent question may generate a fluent, low-entropy response while having stopped attending to the diagram midway through reasoning: confident but visually disengaged. Whether this matters depends on whether the task requires sustained visual reference, and evidence collapse may not manifest only at the final answer but emerge progressively throughout the reasoning trajectory: the confident-but-blind failure mode that is the focus of this work.

If entropy is structurally blind to visual disengagement, we need signals from the visual modality itself. Visual attention to relevant image regions offers a white-box proxy, and cumulative grounding over the full reasoning trajectory may be more informative than any single-point snapshot, since the failure mode is progressive. The central question is \textbf{whether the relationship between visual disengagement and error is consistent across task types}, or whether visual disengagement is itself task-conditional. 

\section{Entropy-Attention Signals: Setup}
\label{sec:framework}
\subsection{Problem Setting}
\label{sec:framework:problem_setting}

A vision-language model receives image $I$ and question $q$, generating response $y = (y_1, \dots, y_T)$ token by token (length $L=T$). In our setting, RVLMs output a reasoning trace enclosed in \texttt{<think>...</think>} followed by a final answer; the \texttt{</think>} token marks the boundary between reasoning and answer spans. Our goal is to estimate $P(\text{correct} \mid I, q, y)$ using signals extractable during or after generation. Correctness is a binary label based on exact match or semantic equivalence with a ground-truth answer.

\subsection{Textual Uncertainty Signal}

Given top-$k$ log-probabilities at each generation step ($k=20$), we compute normalized probabilities and Shannon entropy:
\vspace{-8pt}
\begin{equation}
p_i = \frac{e^{\ell_i}}{\sum_{j=1}^k e^{\ell_j}}, \quad H_t = -\sum_{i=1}^k p_i \log_2 p_i.
\end{equation}
Token-averaged entropy across a span $S$ is $\bar{H}_S = \frac{1}{|S|} \sum_{t \in S} H_t$. Because reasoning-chain dynamics and endpoint confidence may carry different signals, we distinguish two spans: \textbf{Full-response entropy} $\bar{H}_{\text{full}}$, where $S$ spans the entire generation (thinking + answer tokens), and \textbf{Answer-span entropy} $\bar{H}_{\text{ans}}$, where $S$ spans answer tokens only.
\vspace{-8pt}
\subsection{Grounding Layer Selection}
\label{sec:framework:layers}
\vspace{-6pt}
Following \citet{liu2025seeing}, we compute per-layer AUROC on a held-out calibration set (100 Visual-CoT samples with bounding-box annotations), independent of evaluation datasets:
\begingroup
\setlength{\belowdisplayskip}{3pt}
\setlength{\belowdisplayshortskip}{3pt}
\vspace{-8pt}
\begin{equation}
\text{AUROC}^{(\ell)} = \frac{1}{|\mathcal{S}|} \sum_{s \in \mathcal{S}} \text{AUROC}(y^{(s)}_I, \bar{a}^{(\ell,s)}_I).
\end{equation}
\endgroup
Here $\mathcal{S}$ is the calibration set. For each sample $s$, $y^{(s)}_I \in \{0,1\}^{|\mathcal{V}|}$ is a binary mask indicating which visual token positions fall inside the annotated bounding box, and $\bar{a}^{(\ell,s)}_I \in \mathbb{R}^{|\mathcal{V}|}$ is the vector of head-averaged attention weights to visual tokens at layer $\ell$. The inner AUROC measures how well the layer's attention discriminates evidence-region tokens from non-evidence visual tokens; averaging over $\mathcal{S}$ gives the mean evidence localization ability of layer $\ell$.
We select the top-$k$ layers ($k = 6$) as a contiguous block centred on the peak-AUROC layer. Letting $\ell^* = \arg\max_\ell\,\text{AUROC}^{(\ell)}$, the architecture-specific grounding layer set is:
\vspace{-8pt}
\begin{equation}
\mathcal{L}_{\text{VG}} = \bigl\{\ell^* - \lfloor k/2 \rfloor,\;\dots,\;\ell^* + \lceil k/2 \rceil - 1\bigr\}.
\label{eq:lvg}
\end{equation}

\subsection{Visual Attention Mass}
\label{sec:framework:vg}

Let the input contain $m$ visual tokens; let $\mathcal{V}$ denote their index set, so $|\mathcal{V}| = m$. At generation step $t$, head $h$ of layer $\ell$ produces attention weights over all preceding tokens. Averaging over heads and aggregating over architecture-specific grounding layers $\mathcal{L}_{\text{VG}}$, we define total visual attention mass:
\begin{equation}
V_t = \frac{1}{|\mathcal{L}_{\text{VG}}|} \sum_{\ell \in \mathcal{L}_{\text{VG}}} \sum_{v \in \mathcal{V}} \bar{a}^{(\ell)}_{t,v}.
\label{eq:vt}
\end{equation}
Let $\mathcal{B} \subseteq \mathcal{V}$ denote tokens overlapping manually-annotated bounding boxes. Evidence attention mass is defined analogously, restricting the sum to evidence-region tokens:
\begin{equation}
A_{\text{bbox},t} = \frac{1}{|\mathcal{L}_{\text{VG}}|} \sum_{\ell \in \mathcal{L}_{\text{VG}}} \sum_{v \in \mathcal{B}} \bar{a}^{(\ell)}_{t,v}.
\label{eq:abbox}
\end{equation}
Both $V_t$ and $A_{\text{bbox},t}$ are \emph{absolute} mass measures: a model that narrows focus to evidence regions while losing overall visual attention will show decreasing $A_{\text{bbox},t}$ even if the fraction of visual attention on the bbox is unchanged.

\section{Experiments}
\label{sec:experiments}

\subsection{Experimental Setup}
\label{sec:setup}

\textbf{Datasets.}
We evaluate on three benchmarks: \textbf{MathVista}~\citep{lu2024mathvistaevaluatingmathematicalreasoning}, which tests mathematical reasoning over geometric diagrams and algebraic visual expressions; \textbf{HallusionBench}~\citep{guan2024hallusionbenchadvanceddiagnosticsuite}, which uses adversarially modified visual contexts designed to induce hallucinations; and \textbf{MMMU\_Pro}~\citep{yue2025mmmuprorobustmultidisciplinemultimodal}, which poses expert-level STEM questions spanning physics, chemistry, biology, and engineering. From each benchmark we draw a 300-sample stratified subset with manual bounding-box annotation, totaling 900 samples across 2,700 model$\times$example runs.

\textbf{Models.}
We study three models across two architectural families. \textbf{Qwen3-VL-2B-Thinking} (hereafter \textbf{Qwen-2B}) and \textbf{Qwen3-VL-8B-Thinking} (hereafter \textbf{Qwen-8B})~\citep{bai2025qwen3vl} employ the DeepStack architecture with multi-level visual feature injection across the transformer stack~\citep{meng2024deepstack}. \textbf{GLM-4.6V-Flash} (hereafter \textbf{GLM})~\citep{glm2024glm46v} is a single-stream, token-level early-fusion VLM: a ViT-based visual encoder produces patch embeddings that are mapped into the language embedding space by an MLP adapter and injected as image tokens into a 40-layer decoder-only GLM text backbone for autoregressive generation.

\textbf{Annotation.}
Bounding boxes were annotated by two annotators; guideline: box all image regions containing information necessary to answer correctly.

\subsection{Text-Only Entropy Baseline}
\label{sec:entropy_baseline}

Before introducing visual grounding signals, we test whether text-only entropy provides reliable uncertainty estimation. We evaluate both entropy spans defined in \S\ref{sec:framework}: full-generation entropy ($\bar{H}_{\text{full}}$) and answer-time entropy ($\bar{H}_{\text{ans}}$), across all nine model$\times$dataset cells, filtering for normal-stop responses.\footnote{Sample counts after filtering vary by model; full tables in Appendix~\ref{app:entropy_full_tables}.}
We summarize discrimination strength per cell using Cohen's $d$ (correct vs.\ incorrect): negative $d$ means incorrect answers have \emph{higher} entropy than correct ones; positive $d$ means incorrect answers have \emph{lower} entropy.
 
\begin{figure*}[t]
\centering
\includegraphics[width=0.70\textwidth]{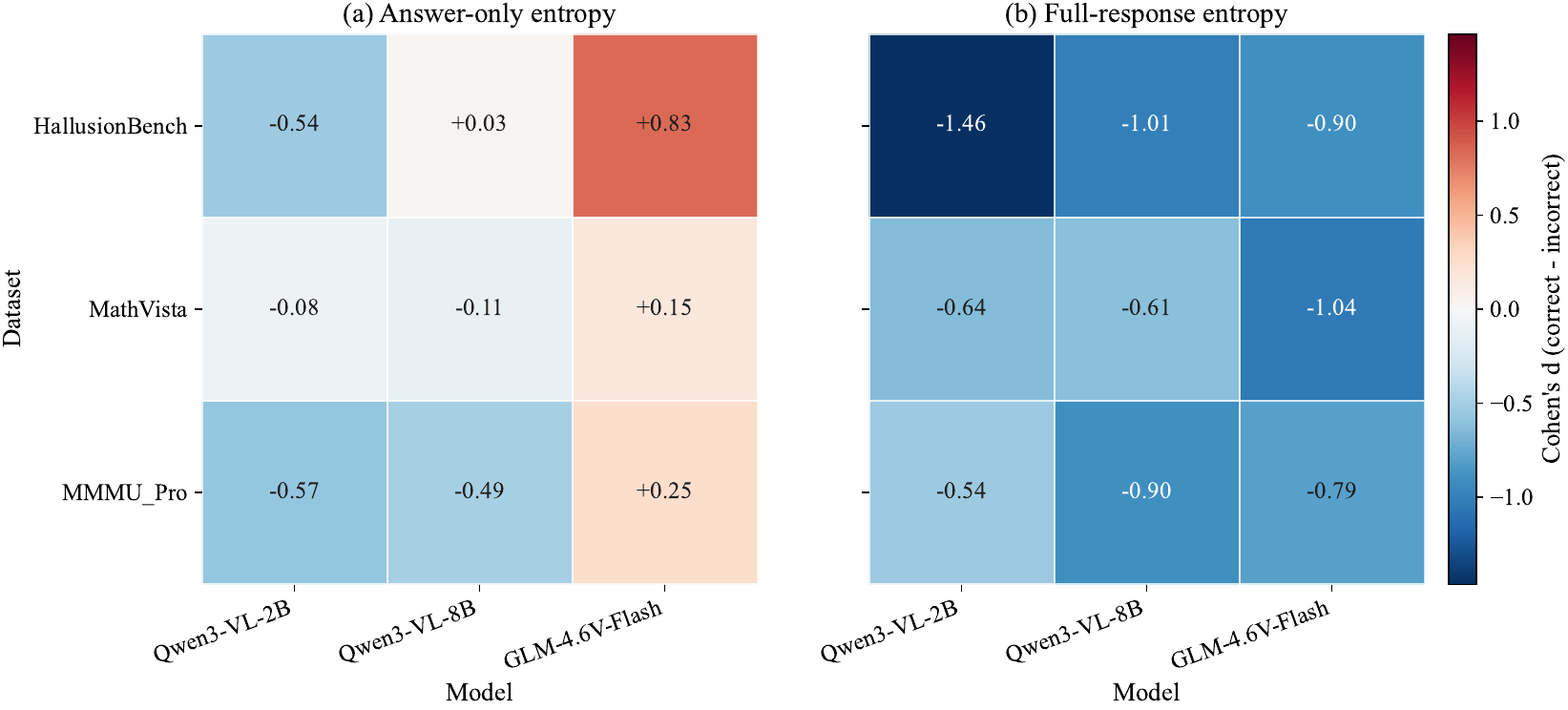}
\caption{\textbf{Entropy discrimination across nine model$\times$dataset cells.} Panel (a) shows answer-time entropy ($\bar{H}_{\text{ans}}$) and panel (b) shows full-generation entropy ($\bar{H}_{\text{full}}$). Values are Cohen's $d$ (correct vs.\ incorrect). For panel (a), significant conventional discrimination appears only in GLM $\times$ HallusionBench. For panel (b), all nine cells are negative (incorrect answers have higher entropy), yielding a more consistent text-only baseline.}
\label{fig:entropy_disc}
\end{figure*}

\textbf{Finding 1: Answer-time entropy ($\bar{H}_{\text{ans}}$) is unstable as a standalone monitor.}
Panel~(a) of Figure~\ref{fig:entropy_disc} shows that significant conventional discrimination (incorrect answers with lower entropy) appears only in GLM $\times$ HallusionBench. Most other cells are either directionally inverted or statistically weak, indicating that answer-time entropy depends strongly on model family and dataset and is not a universal uncertainty signal.

\textbf{Finding 2: Full-generation entropy ($\bar{H}_{\text{full}}$) is the stronger and more consistent text-only baseline.}
Panel~(b) of Figure~\ref{fig:entropy_disc} shows uniformly negative effect sizes across all nine cells (incorrect answers have higher entropy), with all nine significant.\footnote{Exact effect sizes and $p$-values for all nine cells are reported in Appendix Table~\ref{tab:app_entropy_full}.} Effect sizes are larger than answer-time entropy in 8/9 cells, suggesting that aggregating over the full reasoning trace captures uncertainty dynamics that answer-time snapshots miss.

\subsection{Architecture-Dependent Grounding Layers}
\label{sec:layers}

Visual grounding is not uniformly distributed across transformer layers: averaging attention over all layers would dilute the signal from layers that actually encode evidence localization with noise from layers that do not. Before measuring grounding trajectories, we must identify which layers to measure.

\textbf{Experiment.} We compute per-layer AUROC for evidence localization using the procedure defined in \S\ref{sec:framework:layers}: for each layer, we measure how well head-averaged attention weights discriminate bounding-box tokens from non-evidence visual tokens on the held-out calibration set (100 Visual-CoT samples, independent of evaluation datasets). We evaluate all layers for each of the three models.

\begin{figure}[htbp]
\centering
\includegraphics[width=\columnwidth]{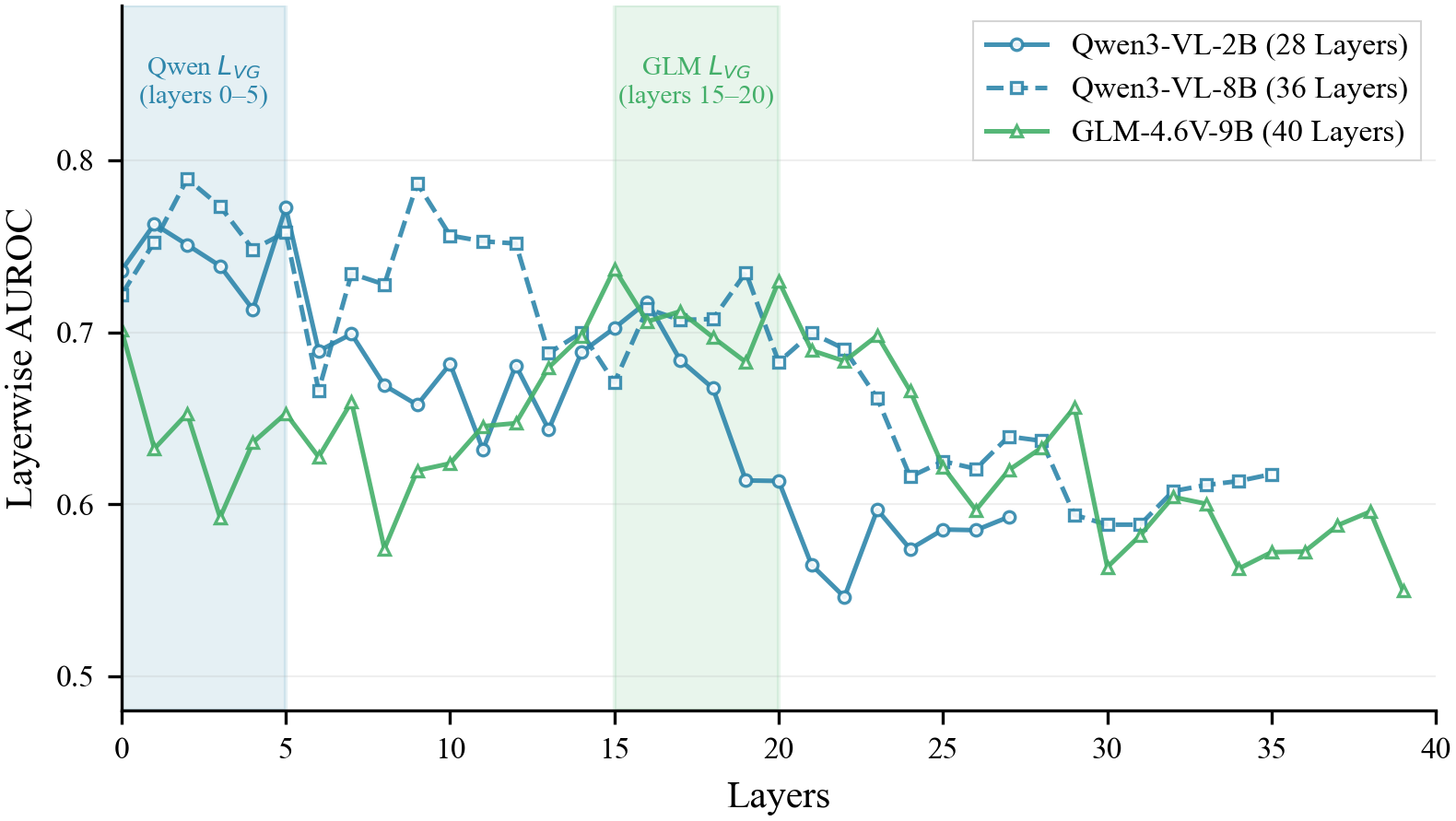}
\caption{\textbf{Architecture-dependent grounding layers.} Layerwise AUROC for evidence localization on a held-out calibration set. Qwen models (DeepStack) peak in early layers (0--5, AUROC $\approx 0.75$--$0.77$). GLM (single-stream early-fusion) peaks in layers 15--20. Grounding layer selection must be architecture-specific.}
\label{fig:layer_auroc}
\end{figure}

\textbf{Results.} Figure~\ref{fig:layer_auroc} shows that grounding layer location differs sharply across architectures. Qwen-2B and Qwen-8B peak in layers 0--5 (AUROC $\approx 0.75$--$0.77$), with weak retention in later layers. GLM peaks in layers 15--20. The pattern is consistent across datasets.

\textbf{Layer selection.} We set $\mathcal{L}_{\text{VG}} = \{0, \ldots, 5\}$ for Qwen-2B and Qwen-8B, and $\mathcal{L}_{\text{VG}} = \{15, \ldots, 20\}$ for GLM, consistent with where visual features enter each architecture's transformer stack~\citep{bai2025qwen3vl,meng2024deepstack}. All subsequent attention metrics are computed over these architecture-specific layers.

\subsection{Evidence Collapse and Error Prediction}
\label{sec:collapse}
\vspace{-2pt}
\textbf{Sampling schedule and canonical positions.}
To characterize grounding dynamics, we sample attention at seven positions spanning the full generation. Five fall within the reasoning trace: \textsc{Start} (first reasoning token), \textsc{Early} (40\% progress), \textsc{Mid} (60\%), \textsc{Late} (80\%), and \textsc{EndThink} (final token before \texttt{</think>}). Two fall within the answer: \textsc{AnsStart} (first answer token) and \textsc{AnsEnd} (last answer token). Formally, let $t_0$ denote the first reasoning token and $t_{\mathrm{think}}$ the last token before \texttt{</think>}, with $\Delta = t_{\mathrm{think}} - t_0$. The five reasoning sampling points are at:
\begingroup
\setlength{\abovedisplayskip}{3pt}
\setlength{\abovedisplayshortskip}{3pt}
\setlength{\belowdisplayskip}{3pt}
\setlength{\belowdisplayshortskip}{3pt}
\begin{equation}
t \in \{t_0,\; t_0 + 0.4\Delta,\; t_0 + 0.6\Delta,\; t_0 + 0.8\Delta,\; t_{\mathrm{think}}\}.
\end{equation}
\endgroup

\textbf{Extraction procedure.}
At each sampled position, we perform a forward pass and compute $V_t$ and $A_{\text{bbox},t}$ (\S\ref{sec:framework:vg}) over $\mathcal{L}_{\text{VG}}$, producing seven-point trajectories per sample. The bounding-box subset $\mathcal{B}$ is determined by mapping visual tokens to a spatial merge grid and selecting those overlapping the annotated evidence region.

\textbf{Trajectory metrics.}
Endpoint metrics capture the residual grounding state at answer time, but two models can arrive at the same $V_{\textsc{AnsEnd}}$ via very different paths: one sustaining engagement throughout reasoning, the other collapsing early. To capture the grounding \emph{process}, we aggregate the five thinking-phase sampling points into trajectory-level features. Treating the five reasoning positions $p_0 = \textsc{Start}, p_1 = \textsc{Early}, p_2 = \textsc{Mid}, p_3 = \textsc{Late}, p_4 = \textsc{EndThink}$ as evenly-spaced ordinal positions $\{0,1,2,3,4\}$, the trapezoidal area under the $V$-trajectory during thinking is:
\begingroup
\setlength{\abovedisplayskip}{3pt}
\setlength{\abovedisplayshortskip}{3pt}
\setlength{\belowdisplayskip}{3pt}
\setlength{\belowdisplayshortskip}{3pt}
\begin{equation}
V_{\text{thinking\_auc}} = \tfrac{1}{2}\sum_{i=0}^{3}(V_{p_i} + V_{p_{i+1}}).
\label{eq:auc}
\end{equation}
\endgroup
Higher $V_{\text{thinking\_auc}}$ indicates sustained visual engagement across the reasoning span; lower values indicate early or progressive visual disengagement.
We also compute a \emph{bbox-weighted variant} by substituting $A_{\text{bbox},t}$ for $V_t$: $A_{\text{bbox},\text{thinking\_auc}}$ measures cumulative evidence-region attention rather than total visual attention. The $V$-based metric requires no annotations; the $A_{\text{bbox}}$-based counterpart requires bounding boxes.

\textbf{Finding 3: Universal visual evidence decay across nine cells.}
Across all nine model$\times$dataset cells, both $V_t$ and $A_{\text{bbox},t}$ decline from \textsc{Start} to \textsc{AnsEnd}. Defining total visual-attention decay as $\Delta V = V_{\textsc{Start}} - V_{\textsc{AnsEnd}}$ and evidence decay as $\Delta A_{\text{bbox}} = A_{\text{bbox},\textsc{Start}} - A_{\text{bbox},\textsc{AnsEnd}}$, per-sample decay is near-universal: $\Delta V > 0$ for 94.3\%--100\% of samples per cell, and $\Delta A_{\text{bbox}} > 0$ for 83.5\%--100\%.\footnote{Complete endpoint statistics for both $\Delta V$ and $\Delta A_{\text{bbox}}$, including per-sample positive-decay fractions, are reported in Appendix~\ref{app:complete_decay_results}.}

\textbf{Finding 4: Decay is evidence-specific, not mere attention redistribution.}
Evidence attention $A_{\text{bbox}}$, restricted to visual tokens overlapping the annotated bounding box, also decays in all nine cells (i.e., $\Delta A_{\text{bbox}} > 0$), with relative loss ranging from 53.3\% (Qwen3-VL-2B $\times$ HallusionBench) to 90.8\% (GLM $\times$ HallusionBench). This rules out a benign explanation in which visual attention decay is harmless redistribution: absolute evidence mass on relevant regions still declines across cells, indicating disengagement from visual evidence rather than stable evidence-preserving reallocation.

\begin{figure*}[t]
\centering
\includegraphics[width=0.96\textwidth]{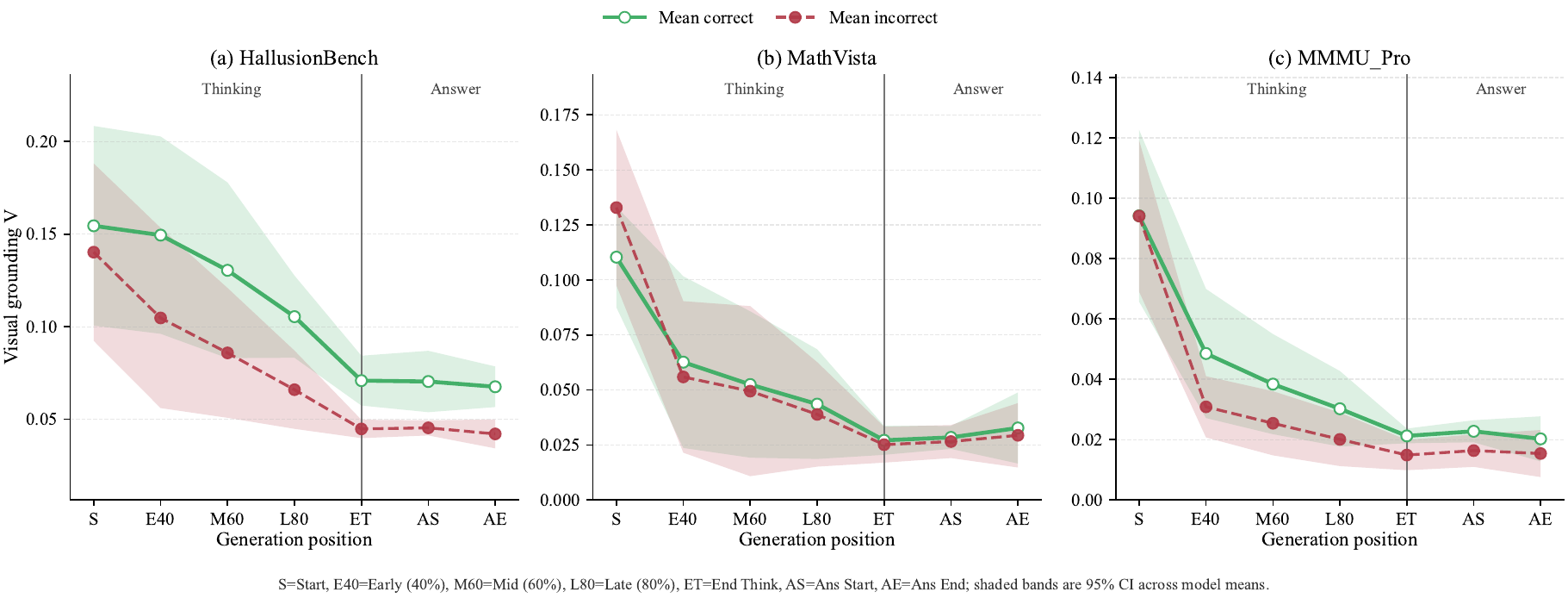}
\caption{\textbf{Three trajectory patterns of visual grounding across seven generation positions.}
Correct (solid) vs.\ incorrect (dashed) with 95\% CI bands.
\emph{Sustained deficit} (HallusionBench): correct maintains higher $V$ throughout.
\emph{Crossover} (MathVista): incorrect starts higher, then crosses below during reasoning.
\emph{Moderate} (MMMU\_Pro): smaller, less consistent separations.
Exact effect sizes and significance tests are reported in Appendix~\ref{app:trajectory_stats}.}
\label{fig:trajectory}
\end{figure*}

\textbf{Finding 5: The grounding--correctness relationship follows dataset-dependent trajectories.}
We track $V_t$ at each of the seven canonical positions and compute Cohen's $d$ per position across model$\times$dataset cells. Figure~\ref{fig:trajectory} shows representative trajectories.
 
Three qualitative patterns recur. \textbf{Sustained grounding deficit} (e.g., HallusionBench, and GLM on MMMU\_Pro): incorrect generations are less grounded than correct ones throughout reasoning and into the answer. \textbf{Crossover} (MathVista): incorrect generations begin more visually engaged, then lose grounding mid-reasoning and fall below correct, which can cancel aggregate trajectory summaries such as $V_{\text{thinking\_auc}}$. \textbf{Moderate effects} (Qwen on MMMU\_Pro): trajectory separations are smaller and less consistent, with substantial overlap between correct and incorrect curves.

Key takeaway: the discriminative grounding signal typically emerges (and often peaks) during the reasoning phase rather than at the answer endpoint, so endpoint-only measurement can underestimate or misstate the pattern.\footnote{See Appendix~\ref{app:trajectory_stats} for effect sizes and significance tests.}

\subsection{Task-Conditional Multimodal Monitoring}
\label{sec:monitoring}

We ask whether visual grounding adds monitoring value beyond text-only entropy, and when it should be trusted. Under cross-dataset transfer, we evaluate a probe ladder and an entropy--vision interaction model.

\textbf{Probe ladder.}
To test whether vision features improve error prediction, we define L2-regularized logistic regression probes that map features to $P(\text{correct})$. We compare nested feature sets of increasing richness: (a)~\emph{entropy-only}: full-generation entropy ($\bar{H}_{\text{full}}$) or answer-time entropy ($\bar{H}_{\text{ans}}$) alone; (b)~\emph{entropy + vision}: adding answer-time visual attention ($V_{\textsc{AnsEnd}}$) and cumulative visual engagement ($V_{\text{thinking\_auc}}$); (c)~\emph{full}: all features plus generation length ($L$) as a control. Features are standardized using training-set statistics only, with per-split regularization strength $C$ tuned via grid search. Probe coefficients reveal which signals matter and in which direction.

\textbf{Transfer protocol.}
A monitor is only useful if it generalizes beyond its training distribution. We evaluate under cross-dataset transfer: train on dataset $D_A$, evaluate on held-out $D_B$. Three datasets with two directions per pair yield six transfers per model and \textbf{18 transfer pairs} across three models.

\textbf{Interaction model.}
The probe ladder tests whether vision features help \emph{linearly}, but the conceptually important question is what happens when entropy says ``confident'' (low $\bar{H}_{\text{full}}$) and vision says ``disengaged'' (low $V_{\text{thinking\_auc}}$) simultaneously. We therefore fit a logistic regression with coefficients $\beta$ for an intercept, entropy main effect, vision main effect, entropy-vision interaction, and length control:
\begin{equation}
\begin{aligned}
\text{logit}(p(\text{correct})) =\;& \beta_0 + \beta_E E_z + \beta_V V_z \\
&+ \beta_{EV}(E_z V_z) + \beta_L L_z
\end{aligned}
\label{eq:interaction}
\end{equation}
where $E_z$, $V_z$, $L_z$ are z-scored full-generation entropy, cumulative visual engagement, and generation length respectively. In the confident-but-blind region ($E_z < 0$, $V_z < 0$), the product $E_z \cdot V_z > 0$, so $\beta_{EV} < 0$ implies a confident-but-blind penalty while $\beta_{EV} > 0$ implies this regime is comparatively benign.

\textbf{Selective prediction.}
Each probe assigns per-example risk $r_i = 1 - P(\text{correct} \mid x_i)$; we defer the riskiest examples to meet a target coverage $\alpha$. We report AUC, coverage-matched AURC, selective risk at fixed coverage, and selective ECE~\citep{naeini2015obtaining}, averaged ($\pm$ s.d.) over $K$ repeats.

\textbf{Validation policy (vision veto).}
To test whether vision can improve entropy-only monitoring as a targeted override rather than linear fusion, we define a two-stage policy (results in Finding~8 below): (1)~entropy-only deferral at target coverage; (2)~among accepted samples, veto the bottom 5\% by cumulative visual engagement $V_{\text{thinking\_auc}}$ (5\% chosen as the highest rate that still allows reliable restoration to 90\% coverage across transfers); (3)~re-loosen entropy threshold to restore target coverage.

\medskip

\textbf{Finding 6: Full-generation entropy is strong; naive fusion fails.}
We compare an entropy-only probe ($\bar{H}_{\text{full}}$) against a naive entropy+vision probe that adds visual grounding features. Under cross-dataset transfer, full-generation entropy is the strongest baseline and consistently outperforms answer-time entropy, while naive fusion does not reliably improve AUC/AURC and often hurts.\footnote{Transfer-level improvement counts and per-model mean AUC deltas are reported in Appendix~\ref{app:fusion_transfer_stats}.}

Entropy is therefore the default monitor, but it has a structural blind spot: it cannot detect visual disengagement.

\begin{figure}[!htbp]
\centering
\includegraphics[width=0.96\columnwidth]{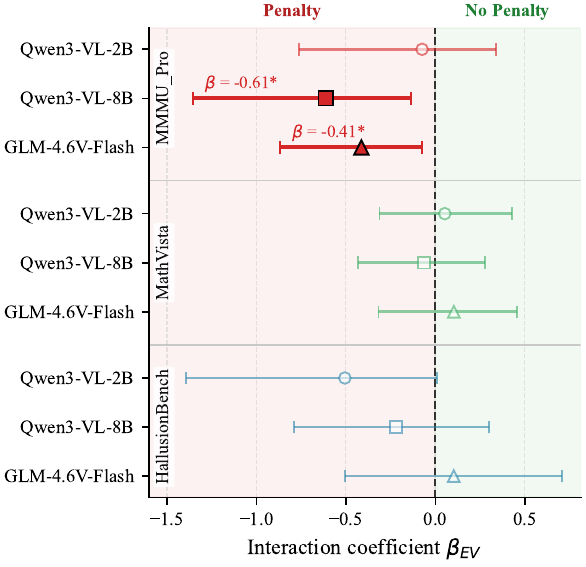}
\caption{\textbf{Task-conditional interaction coefficients ($\beta_{EV}$).} Forest plot with bootstrap 95\% CIs. Negative $\beta_{EV}$ = confident-but-blind penalty. MMMU\_Pro shows significant penalties for GLM ($-0.41$) and Qwen-8B ($-0.61$); MathVista is near zero or positive. This sign flip explains why global linear fusion fails.}
\label{fig:interaction_forest}
\end{figure}

\textbf{Finding 7: The confident-but-blind penalty is task-conditional.}
We fit the interaction model (Eq.~\ref{eq:interaction}) per model $\times$ test dataset cell on $N = 2{,}365$ deduplicated samples. Figure~\ref{fig:interaction_forest} summarizes the interaction coefficients across all nine cells.

\begin{figure*}[!tbp]
\centering
\includegraphics[width=0.86\textwidth]{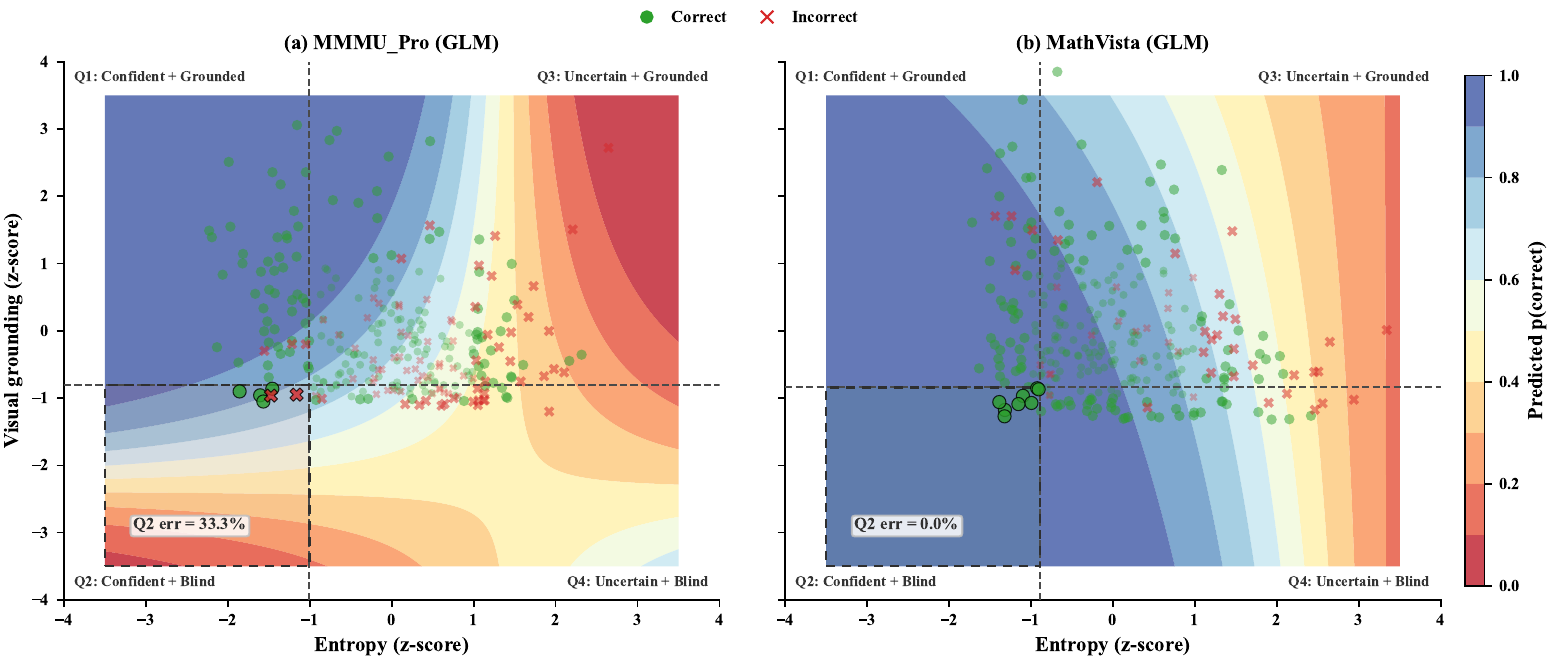}
\caption{\textbf{Regime map of confident-but-blind behavior (GLM representative).} Predicted correctness surface from the continuous interaction model (Eq.~\ref{eq:interaction}) over entropy and visual-grounding z-scores for MMMU\_Pro and MathVista. The Q2 region (confident + blind) is highlighted at $q=0.20$. On MMMU\_Pro, Q2 corresponds to a penalty in predicted correctness; on MathVista, the same region is comparatively safe.}
\label{fig:regime_map}
\end{figure*}

\emph{MMMU\_Pro shows a statistically significant confident-but-blind penalty.} GLM: $\beta_{EV} = -0.414$; Qwen-8B: $\beta_{EV} = -0.613$ (both significant).\footnote{Bootstrap 95\% CIs are reported in Appendix~\ref{app:interaction_cis}.}

\emph{MathVista shows no such penalty.} Interactions are near zero or weakly positive (e.g., GLM: $\beta_{EV} = +0.103$, non-significant), consistent with tasks where symbolic processing correctly replaces sustained visual reference.

\emph{The sign flip between MMMU\_Pro and MathVista is the key finding.} This formalizes why linear fusion fails: a single coefficient cannot accommodate task types where the vision$\to$error relationship has opposite signs. HallusionBench shows weaker and mixed interaction effects due to sparse confident-but-blind samples.\footnote{Additional transfer-level probe-coefficient and length-confound analyses are reported in Appendix~\ref{app:probe_coeff_length}.}

For interpretability, we define quadrants using a binary split at $q = 0.20$ on entropy (confident vs.\ uncertain) and visual engagement (grounded vs.\ blind), a design choice balancing tail focus with usable per-quadrant sample counts: Q1 = confident + grounded, Q2 = confident + blind, Q3 = uncertain + grounded, Q4 = uncertain + blind.
Quadrant analysis corroborates this contrast: the confident+blind region is systematically riskier than confident+grounded on MMMU\_Pro, but comparatively safe on MathVista.\footnote{Exact Q2--Q1 gaps, split-consistency counts, and per-quadrant sample sizes are reported in Appendix~\ref{app:quadrant_stats}.}
Figure~\ref{fig:regime_map} visualizes this contrast as a continuous regime map for GLM.

\textbf{Finding 8: Final validation of the task-conditional structure.}
At 90\% coverage (5\% veto rate), the veto helps where the interaction predicts danger and hurts where it predicts safety (Table~\ref{tab:veto}). Overall, it improves 6/18 transfers and hurts 12/18; the direction is perfectly predicted by the sign of $\beta_{EV}$, validating the task-conditional structure.

\begin{table}[t]
\centering
\small
\caption{\textbf{Vision veto results across transfer directions.} Negative $\Delta$risk = veto helps. The veto helps exactly where the interaction predicts danger (MMMU\_Pro targets) and hurts where it predicts safety (MathVista targets).}
\label{tab:veto}
\begin{tabular}{llcc}
\toprule
Transfer target & Model & $\Delta$risk (pp) & Wins \\
\midrule
MMMU\_Pro & GLM-4.6V-Flash & $-1.94$ & 10/10 \\
MMMU\_Pro & Qwen3-VL-2B & $-0.66$ & 10/10 \\
HallusionBench & GLM-4.6V-Flash & $-1.11$ & 10/10 \\
\midrule
MathVista & GLM-4.6V-Flash & $+1.87$ & 0/10 \\
MathVista & Qwen3-VL-2B & $+1.86$ & 0/10 \\
MathVista & Qwen3-VL-8B & $+1.20$ & 0/10 \\
\bottomrule
\end{tabular}
\end{table}

\vspace{-6pt}
\section{Discussion and Conclusion}
\label{sec:discussion}
\vspace{-6pt}
\textbf{Discussion.}
Extended reasoning can create the same failure mode it is intended to reduce: as generated text grows, text priors strengthen while visual grounding decays. Visual disengagement is not uniformly pathological; it signals danger on tasks requiring sustained visual reference and can be appropriate on tasks that transition to symbolic processing after early visual extraction. This implies that task characterization must precede monitor design, because the same vision signal can indicate opposite risk regimes across tasks. The resulting deployment principle is simple: use full-response entropy as the default monitor and add vision-based checks conditionally when sustained visual grounding is required.

\textbf{Limitations.}
(1) Attention is correlational in this study: we measure attention mass, not causal influence. Establishing causality requires intervention tests such as activation patching.
(2) Model scale is limited to 2B--8B parameters. Generalization to larger RVLMs remains unverified.
(3) Architecture coverage is limited to two model families. Broader architectural diversity may reveal additional grounding dynamics.
(4) Data size is 300 examples per dataset, and normal-stop filtering reduces effective sample size in some cells. This limits precision for cell-level inference.
(5) Extreme-corner analysis at $q=0.20$ uses small per-cell samples (typically $n=2$--$8$), limiting cell-level statistical power. Our primary evidence for task-conditional risk is the continuous interaction model in Eq.~\ref{eq:interaction}, fit over the full sample; quadrant error rates are included only for interpretability and visualization.
(6) Task type is treated as known at analysis time. Zero-shot deployment requires reliable automatic classification of whether sustained visual grounding is needed.

\textbf{Future Work.}
Next steps should move from correlation to intervention by testing whether preserving visual grounding causally improves correctness via activation patching. We also need scale and coverage: 72B+ RVLMs and additional architectures. Practical deployment requires automatic task-type inference that triggers conditional vision monitoring without manual routing.

\textbf{Conclusion.}
We established four findings: evidence collapse is a general property of RVLM generation, full-response entropy is the strongest universal monitor but remains blind to visual disengagement, the error implications of disengagement are task-conditional, and monitoring design must therefore be task-aware rather than globally fused. The deployment question is not whether to add vision, but whether the task requires sustained visual grounding.

\FloatBarrier
\bibliography{references}

\clearpage
\begin{center}
\Large\bfseries Appendix
\end{center}

\appendix

\section{Inference Setup, Prompt Templates, and Answer Parsing}
\label{app:inference_setup}

\subsection{Inference Infrastructure}
\label{app:inference}

Response generation uses vLLM~\citep{kwon2023efficientmemorymanagementlarge} with FlashAttention.
Attention extraction is performed separately: each model is loaded with HuggingFace Transformers (\texttt{attn\_implementation=eager}) and served as a local FastAPI endpoint on a single NVIDIA RTX~3090 (24\,GB).
For each response, we perform seven targeted forward passes at the canonical probe positions defined in \S\ref{sec:collapse} to extract head-averaged attention over $\mathcal{L}_{\text{VG}}$.
Generation hyperparameters are reported in Table~\ref{tab:gen_params} (Appendix~\ref{app:inference}).

\begin{table}[h]
\centering
\caption{Generation hyperparameters (shared across all models and datasets).}
\label{tab:gen_params}
\small
\begin{tabular}{ll}
\toprule
\textbf{Parameter} & \textbf{Value} \\
\midrule
\texttt{max\_tokens}         & 8\,192 \\
\texttt{temperature}         & 0.7 \\
\texttt{top\_p}              & 0.8 \\
\texttt{top\_k}              & 20 \\
\texttt{top\_logprobs}       & 20 \\
\texttt{seed}                & 42 \\
\texttt{system\_prompt}      & \texttt{null} \\
\bottomrule
\end{tabular}
\end{table}

\subsection{Prompt Templates}
\label{app:prompt_templates}

Each dataset uses a fixed prompt template consisting of a \emph{hint} (instructing the desired answer format), the \emph{question}, and, for multiple-choice items, a \emph{choices} block.
The templates are shown below.

\begin{tcolorbox}[
  colback=blue!3, colframe=blue!40!black,
  title={\textbf{MathVista} --- numeric variant},
  fonttitle=\small, fontupper=\small,
  left=3pt, right=3pt, top=1pt, bottom=1pt, boxsep=1pt,
  before skip=3pt, after skip=3pt
]
\texttt{Hint: Please answer the question requiring \{an integer / a floating-point number with one decimal place / a floating-point number with two decimal places\} answer and provide the final value, e.g., \{1, 2, 3 / 1.2, 1.3, 1.4 / 1.23, 1.34, 1.45\}, at the end.}\\[1pt]
\texttt{Question: \{question\}}
\end{tcolorbox}

\begin{tcolorbox}[
  colback=blue!3, colframe=blue!40!black,
  title={\textbf{MathVista} --- multiple-choice variant},
  fonttitle=\small, fontupper=\small,
  left=3pt, right=3pt, top=1pt, bottom=1pt, boxsep=1pt,
  before skip=3pt, after skip=3pt
]
\texttt{Hint: Please answer the question and provide the correct option letter, e.g., A, B, C, D, at the end.}\\[1pt]
\texttt{Question: \{question\}}\\[1pt]
\texttt{Choices: (A) ... (B) ... (C) ... (D) ...}
\end{tcolorbox}

\begin{tcolorbox}[
  colback=green!3, colframe=green!40!black,
  title={\textbf{HallusionBench}},
  fonttitle=\small, fontupper=\small,
  left=3pt, right=3pt, top=1pt, bottom=1pt, boxsep=1pt,
  before skip=3pt, after skip=3pt
]
\texttt{Hint: Please answer the question and provide the correct option letter, e.g., A, B, at the end.}\\[1pt]
\texttt{Question: \{question\}}\\[1pt]
\texttt{Choices: (A) Yes (B) No}
\end{tcolorbox}

\begin{tcolorbox}[
  colback=orange!3, colframe=orange!40!black,
  title={\textbf{MMMU\_Pro}},
  fonttitle=\small, fontupper=\small,
  left=3pt, right=3pt, top=1pt, bottom=1pt, boxsep=1pt,
  before skip=3pt, after skip=3pt
]
\texttt{Hint: Please answer the question and provide the correct option letter, e.g., A, B, C, D, at the end.}\\[1pt]
\texttt{Question: \{question\}}\\[1pt]
\texttt{Choices: (A) ... (B) ... (C) ... (D) ...}
\end{tcolorbox}

\subsection{Output Segmentation and Termination}
\label{app:output_segmentation}

Each assistant response is normalized into \texttt{full\_text}, \texttt{reasoning\_text}, and \texttt{answer\_text} using three parser modes:
\texttt{inline\_think\_tags} (explicit \texttt{<think>...</think>} in content),
\texttt{separate\_reasoning\_field} (reasoning returned in a separate message field), and
\texttt{content\_only} (no explicit reasoning field).
For \texttt{separate\_reasoning\_field}, \texttt{full\_text} is the concatenation of reasoning and answer; otherwise it is the raw content text.

\paragraph{Termination classification.}
We classify each generation into one of three termination types, applied in priority order:
The loop detector uses two complementary repetition measures. The \emph{unique 4-gram ratio} is the number of distinct token 4-grams divided by the total number of 4-grams in a span; a ratio of 1.0 indicates no repeated 4-gram, while values near 0 indicate near-verbatim cycling. The \emph{zlib compression ratio} is the length of the zlib-compressed byte string divided by the uncompressed length; highly repetitive text compresses to a small fraction of its original size, yielding ratios near 0. We require both heuristics to fire jointly to avoid false positives from legitimately structured (but non-looping) reasoning.

\begin{enumerate}[leftmargin=2em,itemsep=2pt]
  \item \textbf{\texttt{max\_tokens\_loop}}: The generation reached \texttt{max\_tokens} \emph{and} exhibits repetition, detected when both the unique 4-gram ratio on the trailing 25\% of tokens falls below 0.5 and the zlib compression ratio of the full text falls below 0.15.
  \item \textbf{\texttt{max\_tokens}}: The generation reached \texttt{max\_tokens} without triggering the loop heuristic above.
  \item \textbf{\texttt{normal\_stop}}: The generation terminated via an EOS token or a successful \texttt{</think>}--to--answer transition.
\end{enumerate}

We mark \texttt{think\_end\_found} if any of the following is true: a token-level \texttt{</think>} position is found, the parsed message structure indicates reasoning has ended, or \texttt{</think>} appears in \texttt{full\_text}.
If termination is \texttt{max\_tokens} or \texttt{max\_tokens\_loop} and \texttt{think\_end\_found} is false, evaluation text is forced to empty.
Otherwise, evaluation text is selected as:
\texttt{answer\_text} for \texttt{separate\_reasoning\_field},
or \texttt{answer\_text} when non-empty (fallback to \texttt{full\_text}) for the other parser modes.

\subsection{Answer Extraction and Correctness Matching}
\label{app:correctness}

All datasets use the same extraction-and-matching procedure on the selected evaluation text.

\paragraph{No-choice path (e.g., MathVista numeric/free-form).}
When \texttt{choices} is empty, the matcher first extracts the last integer from the last non-empty line; if absent, it searches the last 300 characters.
If both prediction and ground truth yield integers, correctness is exact integer equality.
If integer extraction fails, it falls back to normalized string equality.

\paragraph{Choice path (MathVista MCQ, HallusionBench, MMMU\_Pro).}
When \texttt{choices} is non-empty, matching proceeds in order:
\begin{enumerate}[leftmargin=2em,itemsep=2pt]
  \item Extract the last option letter (\texttt{A--D}) from the prediction tail (last 300 characters, then full text as fallback). If found, compare by option index (or directly to letter-form ground truth when applicable).
  \item If no letter is found, choose the option text whose normalized form appears latest in the normalized prediction text.
  \item Final fallback: normalized full-string equality against ground truth.
\end{enumerate}
For HallusionBench, this same choice path is applied with \texttt{choices = [Yes, No]}.

\section{Full Entropy Discrimination Results}
\label{app:entropy_full_tables}

\setlength{\textfloatsep}{8pt plus 1pt minus 2pt}
\setlength{\floatsep}{6pt plus 1pt minus 2pt}
\setlength{\dbltextfloatsep}{8pt plus 1pt minus 2pt}
\setlength{\dblfloatsep}{6pt plus 1pt minus 2pt}
\makeatletter
\setlength{\@fptop}{0pt}
\setlength{\@fpsep}{8pt}
\setlength{\@fpbot}{0pt plus 1fil}
\setlength{\@dblfptop}{0pt}
\setlength{\@dblfpsep}{8pt}
\setlength{\@dblfpbot}{0pt plus 1fil}
\makeatother

This appendix reports the full cell-wise entropy statistics underlying \S\ref{sec:entropy_baseline} and Figure~\ref{fig:entropy_disc}, using the same filtering and effect-size definitions as the main text.

\subsection{Termination Type Breakdown}

Table~\ref{tab:app_termination} reports raw termination outcomes before filtering.

\subsection{Sample Counts After Normal-Stop Filtering}

Table~\ref{tab:app_entropy_counts} shows the effective sample sizes used in all downstream analyses (not just entropy).

\subsection{Full-Response Entropy Discrimination}

Table~\ref{tab:app_entropy_full} gives the exact full-response entropy effects summarized in Figure~\ref{fig:entropy_disc}.

\subsection{Answer-Span Entropy Discrimination}

Table~\ref{tab:app_entropy_answer} reports answer-span entropy for comparison; several cells are non-significant despite consistent inversion in most Qwen cells.

\section{Complete Evidence Decay Results}
\label{app:complete_decay_results}

This appendix provides the full endpoint-decay table referenced in \S\ref{sec:collapse} (Finding~3 and Finding~4), including all nine model$\times$dataset cells and per-sample decay-positive counts.

\subsection{Universal Evidence Decay Tables}

Tables~\ref{tab:decay_v_full} and~\ref{tab:decay_abbox_full} report the full endpoint statistics underlying \S\ref{sec:collapse}, Findings~3 and~4. For each model$\times$dataset cell we record visual attention mass at the first reasoning token and at the last answer token:
\begin{equation}
V_{\text{start}} \coloneqq V_{t_0}, \qquad V_{\text{ans\_end}} \coloneqq V_{t_{\text{ans\_end}}},
\end{equation}
with $V_t$ as defined in Eq.~\ref{eq:vt}. Total visual attention decay is $\Delta V = V_{\text{start}} - V_{\text{ans\_end}}$; evidence attention decay $\Delta A_{\text{bbox}} = A_{\text{bbox,start}} - A_{\text{bbox,ans\_end}}$ is defined analogously using $A_{\text{bbox},t}$ (Eq.~\ref{eq:abbox}). Both quantities are positive when attention decays over the course of generation.

Table~\ref{tab:decay_v_full} reports total visual attention decay. $\Delta V > 0$ in all nine cells, with per-sample decay-positive fractions ranging from 94.3\% to 100\%. Table~\ref{tab:decay_abbox_full} restricts to evidence-region tokens overlapping the annotated bounding box. $\Delta A_{\text{bbox}} > 0$ in all nine cells, with relative loss ranging from 56.8\% (Qwen3-VL-2B $\times$ HallusionBench) to 90.8\% (GLM $\times$ HallusionBench).

\subsection{RVAR Decay Patterns}

Across both Qwen models, mean $\Delta\mathrm{RVAR}$ is negative in every dataset (\S\ref{sec:collapse}), indicating a ``less but smarter'' pattern: residual visual attention becomes relatively more concentrated on relevant regions even as total visual attention drops. Here, the \emph{relevant visual attention ratio} is defined as
\begin{equation}
\mathrm{RVAR}_t = \frac{A_{\text{bbox},t}}{V_t},
\end{equation}
i.e.\ the fraction of total visual attention mass allocated to evidence-region tokens at generation step $t$, and $\Delta\mathrm{RVAR} = \mathrm{RVAR}_{\text{start}} - \mathrm{RVAR}_{\text{ans\_end}}$. For GLM-4.6V-Flash, $\Delta\mathrm{RVAR}$ is mixed and near zero overall (MathVista: $-0.005$, HallusionBench: $+0.038$, MMMU\_Pro: $-0.0002$), so directional interpretation is inconclusive.

This is a secondary observation. The primary quantity in the main text (Finding~3 and Finding~4) is absolute evidence mass, $A_{\text{bbox}} = V \times \mathrm{RVAR}$, which decays across all nine cells (Table~\ref{tab:decay_abbox_full}) regardless of $\mathrm{RVAR}$ direction.

\section{Grounding Trajectory Effect Sizes}
\label{app:trajectory_stats}

This appendix reports the quantitative trajectory discrimination statistics referenced in \S\ref{sec:collapse} (Finding~5). We compute Cohen's $d$ (incorrect vs.\ correct) at each canonical position and use Welch two-sample t-tests for significance.

At \textsc{AnsEnd}, Cohen's $d$ is negative in 9/9 model$\times$dataset cells and significant in 7/9. Peak discrimination occurs during the reasoning phase (\textsc{Start}--\textsc{EndThink}) in 7/9 cells. Endpoint-only measurement can therefore miss substantial signal; the maximum loss is 77.5\% (GLM $\times$ MMMU\_Pro).

The three qualitative trajectory patterns described in \S\ref{sec:collapse} correspond to the following cell-wise statistics:
\begin{itemize}
\item \textbf{Sustained deficit} (HallusionBench $\times$ all models; GLM $\times$ MMMU\_Pro; 4 cells): $V_{\text{thinking\_auc}}$ Cohen's $d$ ranges from $-0.51$ to $-1.05$, all $p < 0.001$.
\item \textbf{Crossover} (MathVista $\times$ all models; 3 cells): at \textsc{Start}, $d = +0.34$ to $+0.41$; $V_{\text{thinking\_auc}}$ is non-significant ($p > 0.35$) due to sign cancellation across the trajectory.
\item \textbf{Moderate} (Qwen $\times$ MMMU\_Pro; 2 cells): Qwen-8B is significant ($d = -0.313$, $p = 0.021$), while Qwen-2B is underpowered ($d = -0.204$, $p = 0.222$, $n = 175$).
\end{itemize}

In aggregate, $V_{\text{thinking\_auc}}$ is significant in 5/9 cells; all three MathVista null results are explained by crossover cancellation.

\section{Naive Fusion Transfer Summary}
\label{app:fusion_transfer_stats}

This appendix reports the summary statistics underlying \S\ref{sec:monitoring}, Finding~6. Table~\ref{tab:app_fusion_transfer_auc} lists all 18 transfer pairs with exact AUC values for entropy-only and entropy+vision probes. Across these 18 pairs, the entropy-plus-vision probe improves AUC in only 4/18 transfers and improves coverage-matched AURC in only 4/18. Mean AUC deltas (vision$-$entropy) are: GLM $= -0.061$, Qwen-2B $= -0.009$, and Qwen-8B $= -0.014$. Full-generation entropy outperforms answer-time entropy in 16/18 transfer pairs.

\section{Interaction Coefficient Confidence Intervals}
\label{app:interaction_cis}

This appendix reports numeric bootstrap 95\% CIs for the key interaction coefficients cited in \S\ref{sec:monitoring} (Finding~7; Figure~\ref{fig:interaction_forest}). MMMU\_Pro: GLM $\beta_{EV}=-0.414$, CI $[-0.868, -0.073]$; Qwen-8B $\beta_{EV}=-0.613$, CI $[-1.356, -0.136]$.

\section{Quadrant Analysis Details}
\label{app:quadrant_stats}

This appendix reports the quadrant summary referenced in \S\ref{sec:monitoring}, Finding~7 (Table~\ref{tab:app_quadrant_q1_q2}). Quadrants are defined by splitting entropy (confident vs.\ uncertain) and visual engagement (grounded vs.\ blind) at $q = 0.20$ (Q1--Q4 as in the main text).
For MMMU\_Pro targets, Q2 (confident + blind) error rates are 15--40 percentage points above Q1 (confident + grounded) for all three models, with 10/10 split consistency in every transfer. For MathVista targets, Q2 error rate is 0\% in all transfers. HallusionBench targets have Q2 sparsity (four transfers with $n_{Q2}=0$), so Q2--Q1 gaps are undefined there. Where Q2 is populated, cell sizes remain small (typically $n = 2$--$8$), so the primary signal is directional consistency rather than per-cell significance.

\section{Probe Coefficient Analysis and Length Confound}
\label{app:probe_coeff_length}

This appendix addresses two methodological questions: what probe coefficients mean, and whether $V_{\text{thinking\_auc}}$ is only a proxy for generation length. We use the same transfer protocol as \S\ref{sec:monitoring}: 18 directed transfers (3 models $\times$ 6 transfers), with $K{=}10$ stratified splits per transfer (180 total fits).

\subsection{Coefficient Sign Consistency Across Transfers}

We report sign consistency as the fraction of fits where a coefficient keeps the same direction. This isolates directional reliability under transfer, independent of absolute magnitude.

Table~\ref{tab:appendix_d_sign_consistency} reports raw sign/non-zero counts only. In addition to the features defined in \S\ref{sec:framework}--\ref{sec:collapse}, the probe ladder includes a \emph{reengagement} feature measuring the recovery of visual attention between the end of reasoning and the answer phase:
\begin{equation}
V_{\text{reengagement}} = V_{\text{ans\_start}} - V_{\text{end\_think}},
\end{equation}
where $V_{\text{ans\_start}}$ and $V_{\text{end\_think}}$ are evaluated at the first answer token and final reasoning token, respectively. Positive values indicate the model re-attends to visual tokens when transitioning to the answer.

The main interpretation is: $V_{\text{thinking\_auc}}$ is the most direction-stable trajectory feature; $V_{\text{ans\_end}}$ is less stable for GLM than for Qwen; $\bar{H}_{\text{full}}$ exhibits model-family direction inversion; $L$ is uniformly negative in the length-only probe; $\mathrm{RVAR}_{\text{ans\_end}}$ is effectively dead for Qwen (all-zero coefficients); $V_{\text{reengagement}}$ is directionally unstable; and $A_{\text{bbox,ans\_end}}$ is stable for GLM but flips sign for Qwen by training dataset.

\subsection{Length Confound Analysis}

This analysis tests a length confound: visual attention may mechanically decay with longer generations, making $V_{\text{thinking\_auc}}$ redundant with $L$. The falsification criterion is that, if $V_{\text{thinking\_auc}}$ were only length, its incremental effect would disappear once length is included.
Table~\ref{tab:appendix_d_length_confound} reports only core quantities: $r_{\cdot}$ are Pearson correlations between $L$ and $V_{\text{thinking\_auc}}$; $N_{\mathrm{PI\text{-}dom}}/N_{\mathrm{transfer}}$ is the number of transfers where $V_{\text{thinking\_auc}}$ has larger $\Delta$NLL permutation importance than $L$ in the joint probe; and $N_{\beta>0}/N_{\mathrm{split}}$ is the number of splits with positive $V_{\text{thinking\_auc}}$ coefficient in that same joint probe.
Interpretation: the correlations are moderately negative rather than near-deterministic, indicating coupling without redundancy. In the joint two-feature probe, $V_{\text{thinking\_auc}}$ dominates in 10/18 transfers, and its coefficient remains positive in 148/180 splits under explicit length control. If $V_{\text{thinking\_auc}}$ were only a length proxy, both results would collapse after conditioning on $L$.

These results are inconsistent with a ``length-only'' explanation. Length and visual engagement are coupled, but not interchangeable, and $V_{\text{thinking\_auc}}$ retains independent predictive semantics under direct length control.

\begin{table*}[t]
\centering
\scriptsize
\setlength{\tabcolsep}{3pt}
\caption{Trajectory discrimination statistics for total visual attention $V_t$. Each entry is Cohen's $d$ (incorrect$-$correct) with Welch $p$-value in parentheses.}
\label{tab:app_trajectory_stats_full}
\resizebox{\textwidth}{!}{%
\begin{tabular}{llccccccccc}
\toprule
Model & Dataset & $n$ & Start & Early & Mid & Late & EndThink & AnsStart & AnsEnd & $V_{\text{thinking\_auc}}$ \\
\midrule
GLM & HallusionBench & 300 & $-0.819$ ($<0.001$) & $-0.488$ ($0.006$) & $-0.729$ ($<0.001$) & $-0.638$ ($<0.001$) & $-0.561$ ($<0.001$) & $-0.510$ ($<0.001$) & $-0.561$ ($<0.001$) & $-0.888$ ($<0.001$) \\
GLM & MathVista & 298 & $+0.389$ ($0.037$) & $-0.132$ ($0.392$) & $+0.061$ ($0.691$) & $-0.096$ ($0.483$) & $+0.001$ ($0.994$) & $+0.038$ ($0.785$) & $-0.172$ ($0.179$) & $+0.007$ ($0.958$) \\
GLM & MMMU\_Pro & 287 & $-0.032$ ($0.827$) & $-0.490$ ($<0.001$) & $-0.352$ ($0.002$) & $-0.401$ ($<0.001$) & $-0.289$ ($0.025$) & $-0.214$ ($0.034$) & $-0.110$ ($0.360$) & $-0.514$ ($<0.001$) \\
Qwen-2B & HallusionBench & 251 & $-0.362$ ($0.033$) & $-0.970$ ($<0.001$) & $-1.046$ ($<0.001$) & $-0.847$ ($<0.001$) & $-1.127$ ($<0.001$) & $-1.161$ ($<0.001$) & $-1.038$ ($<0.001$) & $-1.050$ ($<0.001$) \\
Qwen-2B & MathVista & 239 & $+0.336$ ($0.032$) & $-0.155$ ($0.239$) & $-0.186$ ($0.180$) & $-0.126$ ($0.359$) & $-0.199$ ($0.145$) & $-0.236$ ($0.091$) & $-0.133$ ($0.334$) & $-0.070$ ($0.613$) \\
Qwen-2B & MMMU\_Pro & 175 & $-0.001$ ($0.995$) & $-0.222$ ($0.179$) & $-0.294$ ($0.058$) & $-0.194$ ($0.241$) & $-0.261$ ($0.097$) & $-0.337$ ($0.031$) & $-0.238$ ($0.133$) & $-0.204$ ($0.221$) \\
Qwen-8B & HallusionBench & 299 & $-0.489$ ($0.004$) & $-0.788$ ($<0.001$) & $-0.572$ ($<0.001$) & $-0.680$ ($<0.001$) & $-0.676$ ($<0.001$) & $-0.699$ ($<0.001$) & $-0.810$ ($<0.001$) & $-0.749$ ($<0.001$) \\
Qwen-8B & MathVista & 285 & $+0.406$ ($0.006$) & $-0.253$ ($0.037$) & $-0.278$ ($0.019$) & $-0.357$ ($0.003$) & $-0.230$ ($0.066$) & $-0.228$ ($0.068$) & $-0.258$ ($0.029$) & $-0.110$ ($0.360$) \\
Qwen-8B & MMMU\_Pro & 241 & $-0.034$ ($0.834$) & $-0.375$ ($0.004$) & $-0.403$ ($<0.001$) & $-0.418$ ($<0.001$) & $-0.323$ ($0.013$) & $-0.299$ ($0.024$) & $-0.368$ ($0.002$) & $-0.313$ ($0.021$) \\
\bottomrule
\end{tabular}}
\end{table*}

\begin{table*}[t]
\centering
\scriptsize
\setlength{\tabcolsep}{5pt}
\caption{All 18 transfer pairs for entropy-only vs.\ entropy+vision probes (AUC). $\Delta$AUC is computed as (entropy+vision) $-$ (entropy-only).}
\label{tab:app_fusion_transfer_auc}
\begin{tabular}{lllccc}
\toprule
Model & Train & Test & AUC (Entropy) & AUC (Entropy+Vision) & $\Delta$AUC \\
\midrule
GLM & HallusionBench & MMMU\_Pro & 0.712 & 0.647 & -0.064 \\
GLM & HallusionBench & MathVista & 0.732 & 0.588 & -0.143 \\
GLM & MMMU\_Pro & HallusionBench & 0.755 & 0.770 & +0.015 \\
GLM & MMMU\_Pro & MathVista & 0.732 & 0.686 & -0.045 \\
GLM & MathVista & HallusionBench & 0.755 & 0.676 & -0.079 \\
GLM & MathVista & MMMU\_Pro & 0.712 & 0.661 & -0.051 \\
Qwen-2B & HallusionBench & MMMU\_Pro & 0.649 & 0.651 & +0.002 \\
Qwen-2B & HallusionBench & MathVista & 0.672 & 0.616 & -0.056 \\
Qwen-2B & MMMU\_Pro & HallusionBench & 0.850 & 0.833 & -0.017 \\
Qwen-2B & MMMU\_Pro & MathVista & 0.672 & 0.670 & -0.002 \\
Qwen-2B & MathVista & HallusionBench & 0.850 & 0.809 & -0.041 \\
Qwen-2B & MathVista & MMMU\_Pro & 0.649 & 0.710 & +0.060 \\
Qwen-8B & HallusionBench & MMMU\_Pro & 0.744 & 0.693 & -0.051 \\
Qwen-8B & HallusionBench & MathVista & 0.639 & 0.590 & -0.049 \\
Qwen-8B & MMMU\_Pro & HallusionBench & 0.759 & 0.757 & -0.002 \\
Qwen-8B & MMMU\_Pro & MathVista & 0.639 & 0.681 & +0.043 \\
Qwen-8B & MathVista & HallusionBench & 0.759 & 0.743 & -0.016 \\
Qwen-8B & MathVista & MMMU\_Pro & 0.744 & 0.732 & -0.012 \\
\bottomrule
\end{tabular}
\end{table*}

\begin{table*}[t]
\centering
\scriptsize
\setlength{\tabcolsep}{4pt}
\caption{Quadrant error-rate summary at $q=0.20$ for all 18 directed transfers. Q1 is confident+grounded; Q2 is confident+blind. Entries show mean error rate with mean cell size $n$ in parentheses across 10 splits.}
\label{tab:app_quadrant_q1_q2}
\begin{tabular}{lllcccc}
\toprule
Model & Train & Test & Q1 Error $(n)$ & Q2 Error $(n)$ & Q2$-$Q1 & $\Delta>0$ Splits \\
\midrule
GLM & HallusionBench & MMMU\_Pro & 0.058 (52) & 0.333 (6) & +0.276 & 10/10 \\
GLM & HallusionBench & MathVista & 0.077 (52) & 0.000 (8) & -0.077 & 0/10 \\
GLM & MMMU\_Pro & HallusionBench & 0.033 (60) & -- (0) & -- & 0/10 \\
GLM & MMMU\_Pro & MathVista & 0.077 (52) & 0.000 (8) & -0.077 & 0/10 \\
GLM & MathVista & HallusionBench & 0.033 (60) & -- (0) & -- & 0/10 \\
GLM & MathVista & MMMU\_Pro & 0.058 (52) & 0.333 (6) & +0.276 & 10/10 \\
Qwen-2B & HallusionBench & MMMU\_Pro & 0.000 (29) & 0.400 (5) & +0.400 & 10/10 \\
Qwen-2B & HallusionBench & MathVista & 0.133 (45) & 0.000 (3) & -0.133 & 0/10 \\
Qwen-2B & MMMU\_Pro & HallusionBench & 0.000 (51) & -- (0) & -- & 0/10 \\
Qwen-2B & MMMU\_Pro & MathVista & 0.133 (45) & 0.000 (3) & -0.133 & 0/10 \\
Qwen-2B & MathVista & HallusionBench & 0.000 (51) & -- (0) & -- & 0/10 \\
Qwen-2B & MathVista & MMMU\_Pro & 0.000 (29) & 0.400 (5) & +0.400 & 10/10 \\
Qwen-8B & HallusionBench & MMMU\_Pro & 0.047 (43) & 0.200 (5) & +0.153 & 10/10 \\
Qwen-8B & HallusionBench & MathVista & 0.109 (55) & 0.000 (2) & -0.109 & 0/10 \\
Qwen-8B & MMMU\_Pro & HallusionBench & 0.017 (58) & 0.500 (2) & +0.483 & 10/10 \\
Qwen-8B & MMMU\_Pro & MathVista & 0.109 (55) & 0.000 (2) & -0.109 & 0/10 \\
Qwen-8B & MathVista & HallusionBench & 0.017 (58) & 0.500 (2) & +0.483 & 10/10 \\
Qwen-8B & MathVista & MMMU\_Pro & 0.047 (43) & 0.200 (5) & +0.153 & 10/10 \\
\bottomrule
\end{tabular}
\end{table*}

\begin{table*}[!htbp]
\centering
\small
\setlength{\tabcolsep}{5pt}
\caption{Termination classification across all model$\times$dataset cells. Budget-exhausted combines \texttt{max\_tokens} and \texttt{max\_tokens\_loop} (see \S\ref{app:output_segmentation} for definitions). Qwen3-VL-2B$\times$MMMU\_Pro retains only 58\% of samples after filtering.}
\label{tab:app_termination}
\begin{tabular}{llrrrr}
\toprule
Model & Dataset & Total & normal\_stop & max\_tokens & max\_tokens\_loop \\
\midrule
Qwen3-VL-2B & HallusionBench & 300 & 251 & 0 & 49 \\
Qwen3-VL-2B & MathVista & 300 & 239 & 20 & 41 \\
Qwen3-VL-2B & MMMU\_Pro & 300 & 175 & 28 & 97 \\
\addlinespace
Qwen3-VL-8B & HallusionBench & 300 & 299 & 0 & 1 \\
Qwen3-VL-8B & MathVista & 300 & 285 & 12 & 3 \\
Qwen3-VL-8B & MMMU\_Pro & 300 & 241 & 53 & 6 \\
\addlinespace
GLM-4.6V-Flash & HallusionBench & 300 & 300 & 0 & 0 \\
GLM-4.6V-Flash & MathVista & 300 & 298 & 1 & 1 \\
GLM-4.6V-Flash & MMMU\_Pro & 300 & 287 & 11 & 2 \\
\bottomrule
\end{tabular}
\end{table*}

\begin{table*}[!htbp]
\centering
\small
\setlength{\tabcolsep}{5pt}
\caption{Sample counts after normal-stop filtering.}
\label{tab:app_entropy_counts}
\begin{tabular}{llrrrr}
\toprule
Model & Dataset & Total Samples & Normal-Stop Samples & Correct (normal-stop) & Incorrect (normal-stop) \\
\midrule
Qwen3-VL-2B & HallusionBench & 300 & 251 & 213 & 38 \\
Qwen3-VL-2B & MathVista & 300 & 239 & 188 & 51 \\
Qwen3-VL-2B & MMMU\_Pro & 300 & 175 & 120 & 55 \\
\addlinespace
Qwen3-VL-8B & HallusionBench & 300 & 299 & 265 & 34 \\
Qwen3-VL-8B & MathVista & 300 & 285 & 223 & 62 \\
Qwen3-VL-8B & MMMU\_Pro & 300 & 241 & 183 & 58 \\
\addlinespace
GLM-4.6V-Flash & HallusionBench & 300 & 300 & 264 & 36 \\
GLM-4.6V-Flash & MathVista & 300 & 298 & 251 & 47 \\
GLM-4.6V-Flash & MMMU\_Pro & 300 & 287 & 200 & 87 \\
\bottomrule
\end{tabular}
\end{table*}

\begin{table*}[!htbp]
\centering
\small
\setlength{\tabcolsep}{5pt}
\caption{Full-response entropy discrimination ($\bar{H}_{\text{full}}$).}
\label{tab:app_entropy_full}
\begin{tabular}{llrrrrr}
\toprule
Model & Dataset & $N$ & Mean $\bar{H}_{\text{full}}$ (correct) & Mean $\bar{H}_{\text{full}}$ (incorrect) & Cohen's $d$ & $p$-value \\
\midrule
Qwen3-VL-2B & HallusionBench & 251 & 0.393 & 0.687 & -1.46 & $<10^{-12}$ \\
Qwen3-VL-2B & MathVista & 239 & 0.462 & 0.576 & -0.64 & $<10^{-3}$ \\
Qwen3-VL-2B & MMMU\_Pro & 175 & 0.582 & 0.717 & -0.54 & $<10^{-3}$ \\
\addlinespace
Qwen3-VL-8B & HallusionBench & 299 & 0.305 & 0.446 & -1.01 & $<10^{-5}$ \\
Qwen3-VL-8B & MathVista & 285 & 0.350 & 0.438 & -0.61 & $<10^{-3}$ \\
Qwen3-VL-8B & MMMU\_Pro & 241 & 0.440 & 0.586 & -0.90 & $<10^{-8}$ \\
\addlinespace
GLM-4.6V-Flash & HallusionBench & 300 & 0.369 & 0.537 & -0.90 & $<10^{-6}$ \\
GLM-4.6V-Flash & MathVista & 298 & 0.398 & 0.591 & -1.04 & $<10^{-5}$ \\
GLM-4.6V-Flash & MMMU\_Pro & 287 & 0.534 & 0.697 & -0.79 & $<10^{-9}$ \\
\bottomrule
\end{tabular}

\begin{minipage}{\textwidth}
\footnotesize
\emph{Note.} Negative $d$ indicates incorrect answers have higher entropy (inverted discrimination); positive $d$ indicates conventional discrimination.
\end{minipage}
\end{table*}

\begin{table*}[!htbp]
\centering
\small
\setlength{\tabcolsep}{5pt}
\caption{Answer-span entropy discrimination ($\bar{H}_{\text{ans}}$). Strong effects ($|d|>0.5$) are bolded.}
\label{tab:app_entropy_answer}
\begin{tabular}{llrrrrr}
\toprule
Model & Dataset & $N$ & Mean $\bar{H}_{\text{ans}}$ (correct) & Mean $\bar{H}_{\text{ans}}$ (incorrect) & Cohen's $d$ & $p$-value \\
\midrule
Qwen3-VL-2B & HallusionBench & 251 & 0.137 & 0.249 & \textbf{-0.54} & 0.02 \\
Qwen3-VL-2B & MathVista & 239 & 0.263 & 0.278 & -0.08 & 0.7 \\
Qwen3-VL-2B & MMMU\_Pro & 175 & 0.289 & 0.441 & \textbf{-0.57} & 0.003 \\
\addlinespace
Qwen3-VL-8B & HallusionBench & 299 & 0.395 & 0.389 & +0.03 & 0.9 \\
Qwen3-VL-8B & MathVista & 285 & 0.266 & 0.285 & -0.11 & 0.4 \\
Qwen3-VL-8B & MMMU\_Pro & 241 & 0.321 & 0.419 & -0.49 & $<10^{-3}$ \\
\addlinespace
GLM-4.6V-Flash & HallusionBench & 300 & 0.096 & 0.055 & \textbf{+0.83} & $<10^{-6}$ \\
GLM-4.6V-Flash & MathVista & 298 & 0.113 & 0.101 & +0.15 & 0.3 \\
GLM-4.6V-Flash & MMMU\_Pro & 287 & 0.049 & 0.035 & +0.25 & 0.06 \\
\bottomrule
\end{tabular}

\begin{minipage}{\textwidth}
\footnotesize
\emph{Note.} Negative $d$ indicates incorrect answers have higher entropy (inverted discrimination); positive $d$ indicates conventional discrimination.
\end{minipage}
\end{table*}

\begin{table*}[!htbp]
\centering
\normalsize
\setlength{\tabcolsep}{5pt}
\caption{\textbf{Visual attention decay ($V$) across all nine model$\times$dataset cells.} $N$ denotes normal-stop samples.}
\label{tab:decay_v_full}
\resizebox{\textwidth}{!}{%
\begin{tabular}{llrccccl}
\toprule
Model & Dataset & $N$ (normal-stop) & $V_{\text{start}}$ & $V_{\text{ans\_end}}$ & $\Delta V$ (absolute) & $\Delta V$ (\%) & Samples with $\Delta V>0$ \\
\midrule
Qwen3-VL-2B & MathVista & 239 & 0.095 & 0.017 & 0.078 & 82.5 & 239/239 (100.0\%) \\
Qwen3-VL-2B & HallusionBench & 251 & 0.125 & 0.044 & 0.081 & 64.6 & 251/251 (100.0\%) \\
Qwen3-VL-2B & MMMU\_Pro & 175 & 0.082 & 0.012 & 0.070 & 85.4 & 175/175 (100.0\%) \\
\addlinespace
Qwen3-VL-8B & MathVista & 285 & 0.115 & 0.015 & 0.100 & 86.8 & 285/285 (100.0\%) \\
Qwen3-VL-8B & HallusionBench & 299 & 0.172 & 0.038 & 0.134 & 78.0 & 299/299 (100.0\%) \\
Qwen3-VL-8B & MMMU\_Pro & 241 & 0.099 & 0.009 & 0.090 & 90.9 & 241/241 (100.0\%) \\
\addlinespace
GLM-4.6V-Flash & MathVista & 298 & 0.133 & 0.043 & 0.090 & 67.5 & 281/298 (94.3\%) \\
GLM-4.6V-Flash & HallusionBench & 300 & 0.141 & 0.017 & 0.125 & 88.2 & 299/300 (99.7\%) \\
GLM-4.6V-Flash & MMMU\_Pro & 287 & 0.091 & 0.016 & 0.075 & 82.4 & 284/287 (99.0\%) \\
\bottomrule
\end{tabular}
}
\end{table*}

\begin{table*}[!htbp]
\centering
\normalsize
\setlength{\tabcolsep}{3pt}
\caption{\textbf{Evidence attention decay ($A_{\text{bbox}}$) across all nine model$\times$dataset cells.} $N$ denotes normal-stop samples. $A_{\text{bbox}}$ decay-positive counts use valid bbox samples only (denominator shown in the last column).}
\label{tab:decay_abbox_full}
\resizebox{\textwidth}{!}{%
\begin{tabular}{llrccccl}
\toprule
Model & Dataset & $N$ (normal-stop) & $A_{\text{bbox,start}}$ & $A_{\text{bbox,ans\_end}}$ & $\Delta A_{\text{bbox}}$ (absolute) & $\Delta A_{\text{bbox}}$ (\%) & Samples with $\Delta A_{\text{bbox}}>0$ \\
\midrule
Qwen3-VL-2B & MathVista & 239 & 0.017 & 0.003 & 0.014 & 82.0 & 191/191 (100.0\%) \\
Qwen3-VL-2B & HallusionBench & 251 & 0.015 & 0.007 & 0.009 & 56.8 & 250/251 (99.6\%) \\
Qwen3-VL-2B & MMMU\_Pro & 175 & 0.019 & 0.003 & 0.016 & 84.2 & 116/116 (100.0\%) \\
\addlinespace
Qwen3-VL-8B & MathVista & 285 & 0.024 & 0.003 & 0.020 & 85.7 & 220/223 (98.7\%) \\
Qwen3-VL-8B & HallusionBench & 299 & 0.024 & 0.005 & 0.019 & 77.8 & 299/299 (100.0\%) \\
Qwen3-VL-8B & MMMU\_Pro & 241 & 0.029 & 0.003 & 0.026 & 89.7 & 164/164 (100.0\%) \\
\addlinespace
GLM-4.6V-Flash & MathVista & 298 & 0.039 & 0.012 & 0.027 & 68.7 & 193/231 (83.5\%) \\
GLM-4.6V-Flash & HallusionBench & 300 & 0.028 & 0.003 & 0.026 & 90.8 & 300/300 (100.0\%) \\
GLM-4.6V-Flash & MMMU\_Pro & 287 & 0.033 & 0.007 & 0.026 & 78.8 & 178/184 (96.7\%) \\
\bottomrule
\end{tabular}
}
\end{table*}

\begin{table*}[!htbp]
\centering
\small
\setlength{\tabcolsep}{6pt}
\caption{Coefficient-sign consistency across transfer evaluation. Unless noted otherwise, entries report positive-sign fits out of 60 per model (6 transfers $\times$ 10 splits).}
\label{tab:appendix_d_sign_consistency}
\begin{tabular}{lccc}
\toprule
Feature & GLM & Qwen-2B & Qwen-8B \\
\midrule
$V_{\text{thinking\_auc}}$ & $+54/-6$ & $+50/-10$ & $+58/-2$ \\
$V_{\text{ans\_end}}$$^{\dagger}$ & $+10/-8$ & $+16/-2$ & $+18/-0$ \\
$\bar{H}_{\text{full}}$$^{\ddagger}$ & $+6/-0$ & $+2/-4$ & $+0/-6$ \\
$L$$^{\S}$ & $+0/-18$ & $+0/-18$ & $+0/-18$ \\
$\mathrm{RVAR}_{\text{ans\_end}}$$^{\dagger}$ & $12/18$ & $0/18$ & $0/18$ \\
$V_{\text{reengagement}}$ & $+22/-38$ & $+58/-2$ & $+8/-52$ \\
$A_{\text{bbox,ans\_end}}$$^{\P}$ & $+6/-0$ & $+4/-2$ & $+4/-2$ \\
\bottomrule
\end{tabular}
\begin{minipage}{\textwidth}
\footnotesize
\emph{Notes.} Entries are positive/negative sign counts, except $\mathrm{RVAR}_{\text{ans\_end}}$ which reports non-zero/total counts.
$^{\dagger}$18 directed model-transfer cells per model.
$^{\ddagger}$$\bar{H}_{\text{full}}$: 6 directed transfer cells per model (full probe).
$^{\S}$$L$: length-only probe, 18 directed model-transfer cells per model.
$^{\P}$$A_{\text{bbox,ans\_end}}$: 6 directed transfer cells per model (headline entropy+vision probe).
\end{minipage}
\end{table*}

\begin{table*}[!htbp]
\centering
\small
\setlength{\tabcolsep}{8pt}
\caption{Length-confound diagnostics for $V_{\text{thinking\_auc}}$.}
\label{tab:appendix_d_length_confound}
\begin{tabular}{lr}
\toprule
Metric & Value \\
\midrule
$r_{\text{overall}}$ & $-0.43$ \\
$r_{\text{GLM}}$ & $-0.41$ \\
$r_{\text{Qwen-2B}}$ & $-0.59$ \\
$r_{\text{Qwen-8B}}$ & $-0.55$ \\
$N_{\mathrm{PI\text{-}dom}}/N_{\mathrm{transfer}}$ & $10/18$ \\
$N_{\beta>0}/N_{\mathrm{split}}$ & $148/180$ \\
\bottomrule
\end{tabular}
\end{table*}

\end{document}